\documentclass[conference]{IEEEtran}
\IEEEoverridecommandlockouts

\usepackage{amsmath,amssymb,amsfonts}
\usepackage{algorithmic}
\usepackage{graphicx}
\usepackage{textcomp}
\usepackage{xcolor}
\def\BibTeX{{\rm B\kern-.05em{\sc i\kern-.025em b}\kern-.08em
    T\kern-.1667em\lower.7ex\hbox{E}\kern-.125emX}}

\usepackage{times}
\usepackage{latexsym}
\usepackage[T1]{fontenc}

\usepackage[utf8]{inputenc}

\usepackage{microtype}

\usepackage{inconsolata}

\usepackage{graphicx}

\usepackage{booktabs}
\usepackage{multirow}

\usepackage{url}

\usepackage{xspace}
\usepackage{enumitem}

\usepackage{hyperref}
\usepackage[capitalize]{cleveref}

\usepackage{bm}

\usepackage[style=ieee,maxbibnames=5]{biblatex}
\newcommand{\gemma}{Gemma\xspace} %
\newcommand{\qwen}{Qwen-2.5\xspace} %
\newcommand{\llama}{Llama-3.1\xspace} %
\newcommand{\micphi}{Phi-4\xspace}
\newcommand{\github}{\url{https://github.com/GianCarloMilanese/llm_pipeline_wi-iat}\xspace}

\newcommand{\nltk}{NLTK\xspace}
\newcommand{\spacy}{spaCy\xspace}

\newcommand\score[1]{\textit{#1}}
\newcommand{\sw}  {\score{SW}\xspace}
\newcommand{\swl}  {\score{SW+L}\xspace}
\newcommand{\sws}  {\score{SW+S}\xspace}
\newcommand{\nsw} {\score{NSW}\xspace}
\newcommand{\ls}  {\score{L}\xspace}
\renewcommand{\ss}{\score{S}\xspace}

\newcommand{\ul}[1]{\underline{#1}}

\newcommand{\bestinline}{\bm{\dagger}}
\newcommand{\best}{$^{\bestinline}$}

\begin{document}

\title{Investigating Large Language Models' Linguistic Abilities for Text Preprocessing}

\author{\IEEEauthorblockN{Marco Braga}
\IEEEauthorblockA{\textit{University of Milano-Bicocca} \\
  Milan, Italy\\
  \textit{Politecnico di Torino} \\
  Turin, Italy \\
m.braga@campus.unimib.it}
\and
\IEEEauthorblockN{Gian Carlo Milanese}
\IEEEauthorblockA{\textit{University of Milano-Bicocca} \\
  Milan, Italy\\
giancarlo.milanese@unimib.it}
\and
\IEEEauthorblockN{Gabriella Pasi}
\IEEEauthorblockA{\textit{University of Milano-Bicocca} \\
  Milan, Italy\\
gabriella.pasi@unimib.it}
}

\maketitle

\begin{abstract}
Text preprocessing is a
fundamental component of
Natural Language Processing,
involving techniques such as stopword removal, stemming, and lemmatization
to prepare text as input for further processing and analysis.
Despite the context-dependent nature of the above techniques,
traditional
methods
usually ignore contextual information.
In this paper, we investigate the idea of using
Large Language Models (LLMs)
to perform various preprocessing tasks,
due to their ability to take context into account
without requiring extensive
language-specific annotated resources.
Through a comprehensive evaluation on web-sourced data, we compare LLM-based preprocessing
(specifically stopword removal, lemmatization and stemming) to
traditional algorithms across multiple text classification tasks
in six European languages.
Our analysis indicates that LLMs are capable of replicating
  traditional stopword removal, lemmatization, and stemming methods with
  accuracies reaching 97\%, 82\%, and 74\%, respectively.
Additionally,
we show that ML algorithms trained on texts
preprocessed by LLMs achieve an improvement of up to $6\%$
with respect to the $F_1$ measure compared to traditional techniques.
Our code, prompts, and results are publicly available at \github.
\end{abstract}


\section{Introduction}
\label{sec:intro}

Text preprocessing is a fundamental step in Natural Language Processing
(NLP), involving techniques such as stopword removal, stemming, and
lemmatization to standardize text for further processing or downstream
tasks, including input preparation for Machine Learning (ML) algorithms.
By reducing text to its basic features,
text preprocessing decreases the computational cost of the subsequent
processing and mitigates noise and irrelevant information
\cite{hofstatter2020}.
The choice of preprocessing strategy can significantly impact
downstream performance, sometimes enabling even simple models to outperform
complex transformer-based architectures \cite{Siino2024}.

Several preprocessing techniques, such as stopword removal
and lemmatization, are inherently context-dependent.
Indeed, what qualifies as a stopword often varies across tasks and domains,
as each is characterized by a
distinct
word distribution.
Additionally, the context of a text is crucial in determining whether a
word should be treated as a stopword \cite{hofstatter2020}.
Similarly, in lemmatization,
the part of speech of a word often determines how it
should be processed:
for instance, the word ``saw'' may be reduced to either ``see'' or
``saw'' depending on whether it functions as a verb or a noun.
Moreover,
the broader context of a document is also valuable for accurate
lemmatization, as word meanings
can shift significantly based on the subject matter. For example, the
noun ``leaves'' could be lemmatized to ``leaf'' in a document about botany,
but it would be lemmatized to ``leave'' in a text about employee
absences.
As the above examples show, text preprocessing depends not only on the task at
hand or on the part of speech of a word, but also on the broader context
of a sentence or document.
However, traditional preprocessing techniques rely only marginally on
contextual information.
Indeed, they
often make use of predefined
stopwords lists and stemming or lemmatization rules
that
overlook domain-specific information.
These issues highlight the need for techniques that enable a more
context-sensitive
text preprocessing.
To fill this
gap, we investigate the ability of pre-trained Large Language Models (LLMs)
to preprocess a text.
Due to their ability to
take the linguistic context into account
\cite{radford2019language,brown2020language,schick2021exploiting,plaza2023leveraging}
without requiring
extensive language-specific
annotated resources, we hypothesize that LLMs can dynamically detect stopwords, lemmas
and stems based on the input document, context and task.
Although prior work by \cite{Sigir_stem} has examined the role of LLMs in
stemming within information retrieval pipelines, %
the reported study primarily focuses on retrieval effectiveness rather than
the quality of preprocessing itself.
In this paper, we thoroughly investigate the ability of LLMs to perform
text preprocessing, guided by the following research questions:
    \textbf{(RQ1)} How effectively can pre-trained LLMs perform stopword removal,
    stemming, and lemmatization, and how does their performance vary across
    different languages?
    \textbf{(RQ2)} Does the use of LLMs for text preprocessing, as opposed to traditional
    methods, improve the performance on downstream tasks?

\begin{figure*}
    \centering
    \includegraphics[width=\textwidth]{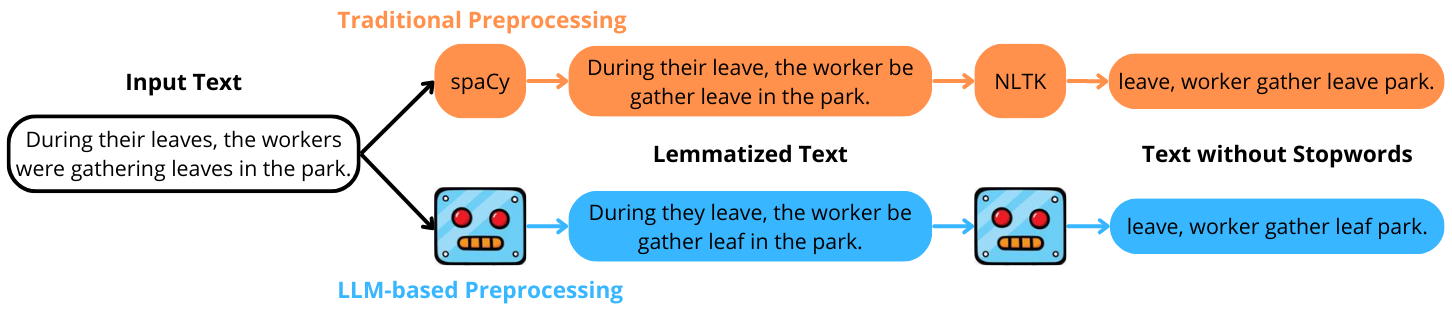}
    \caption{
      Example of traditional vs.~LLM-based text preprocessing.
      In this case, the LLM correctly disambiguates the word ``leaves'',
    distinguishing between employee absences and foliage in its two
  occurrences, and applies lemmatization accordingly.}
    \label{tab:intro_example}
\end{figure*}

To address these questions, we employ recent LLMs, namely
Gemma-2  \cite{team2024gemma} and -3 \cite{team2025gemma},
LLama-3 \cite{dubey2024llama},
Phi-4 \cite{abdin2024phi}, and
Qwen-2.5 \cite{qwen2024},
and we instruct them to
remove stopwords, and to lemmatize or stem a document given a few examples and
the task we are tackling. Furthermore, to comparatively evaluate the
effectiveness of
LLM-based preprocessing,
we train three different
ML-based classification models by using
data preprocessed by the LLMs. In detail, we preprocess data from
multiple sources, e.g., English Social Networks and News, to study the
impact of LLM preprocessing on both dirty and clean data. Furthermore, we
analyse the preprocessing of both English-only and multilingual LLMs on
six European Languages, i.e. English, French, German, Italian, Portuguese
and Spanish, to understand the impact of LLM preprocessing on different
languages.
Our analysis shows that LLMs can replicate traditional stopword
  removal, lemmatization, and stemming methods with accuracies of up to
  97\%, 82\%, and 74\%, respectively.
Furthermore,
we note that ML algorithms trained on texts preprocessed by LLMs achieve an
improvement of up to $6\%$
with respect to the $F_1$ measure compared to
traditional techniques.
The paper is
organised as follows:
after discussing the related works in \cref{sec:related},
\cref{sec:method}
presents in detail the methodology of our analysis,
while in \cref{sec:experimental} we describe the
experimental setup used. Then, \cref{sec:results} discusses the
results of our evaluation, and \cref{sec:limitations} addresses the
limitations of using LLMs for preprocessing.
Finally,
\cref{sec:conclusions_and_further_research} concludes the study and
outlines future research directions.

\section{Related Works}
\label{sec:related}

LLMs have achieved state-of-the-art performance across a wide range of
tasks and research fields \cite{Survey_llm}. They are particularly
effective in few-shot settings, where they can be applied to unseen tasks
or domains without requiring additional supervised fine-tuning
\cite{NEURIPS2020_1457c0d6,agrawal-etal-2022-large,wang2024zero}, which
demands a large amount of labelled data that are not always available \cite{thakur2beir}.
The
relationships
between preprocessing
operations, such as lemmatization and stopword removal, and the context of
input texts has been studied for a long time \cite{dolamic2010stopword,zaman2011evaluation,hofstatter2020,toporkov-agerri-2024-role}.
For instance, \cite{hofstatter2020} shows how to define context-specific stopwords within an information retrieval pipeline: removing
context-specific stopwords achieves higher performance compared to removing
them from a predefined list.
Recently, LLMs have been applied in a few-shot scenario for stemming queries and documents in an
information retrieval pipeline \cite{Sigir_stem}. The authors found that,
while
LLM-based stemming alone does not improve retrieval performance, using LLMs
to identify named entities that should not be stemmed 
leads to significant improvements.
However, no prior work has conducted a comprehensive
analysis of LLMs for text preprocessing - including stopword removal,
lemmatization, and stemming - by comparing their outputs to those produced by traditional
methods, and by assessing their impact on text classification. The study presented in this paper aims to fill that gap.

\section{Methodology}
\label{sec:method}

Our investigation into the text preprocessing capabilities of LLMs
  involves defining prompts that guide these models through each
preprocessing task.
In detail, the LLMs are provided with
(i) a formal description of the target preprocessing operation,
(ii) a few examples of how it should be performed,
(iii) the text to be preprocessed,
(iv) the language of the text,
and %
(v) the context of the downstream task that we are addressing.
The text is directly fed into the LLMs, which output the corresponding
preprocessed version.
Note that our
methodology relies on in-context learning,
as we provide the LLMs with a few
examples of stopwords, lemmas and stems inside the prompt.
With respect to stopword removal, we additionally instruct the LLMs to
retain certain context- and task-specific words that are generally
considered stopwords.
For example, in the sentiment analysis tasks, the LLMs are instructed to
keep the word ``not'' in the text, due to its key role in determining
polarity.
Additionally,
we evaluate our method across multiple languages -- English, French,
German, Italian, Portuguese and Spanish -- to investigate cross-linguistic
performance.
For each non-English
language, we perform experiments with the same prompts written both in English and in that
specific language to assess whether using the native language offers
additional contextual benefits.
To address RQ1, we compare the output of each LLM with the one produced
from the same text preprocessed by using traditional methods.
Specifically, these include
removing words from a predefined stopwords list, applying stemming
algorithms such as Porter \cite{porter1980algorithm}, Lancaster
\cite{PaiceLancaster1990}, and Snowball \cite{porter2001snowball}, and
utilizing off-the-shelf implementations of
rule-based %
or edit tree %
lemmatizers \cite{muller2024}.
With respect to RQ2, we analyze the impact of preprocessing on downstream classification tasks.
We represent the preprocessed texts
as bag-of-words
with TF-IDF \cite{AIZAWA200345}.
Then, we train three well-known ML algorithms,
i.e. Decision Tree \cite{de2013decision}, Logistic Regression
\cite{nick2007logistic}, and Naive Bayes \cite{webb2010naive}.
To assess the overall impact of text preprocessing across
the previously mentioned ML
algorithms, we average the single models' performances.

\section{Experimental Setup}
\label{sec:experimental}

\begin{table*}[ht!]
  \centering
  \resizebox{\textwidth}{!}{
    \small
  \begin{tabular}{p{\textwidth}}
    \toprule
    \multicolumn{1}{c}{\textbf{Stopword removal}}\\
    You specialize in removing stopwords from text. Stopwords are
      words that are not relevant for processing a text. Stopwords
      typically include articles, prepositions, pronouns, and auxiliary
      verbs. For example, the words `is', `are', `being', `you', `me',
      `the', `an', `and', `I', `which', `that', `have', `by', `for' and
      their alternative forms are usually considered stopwords. Note that
      whether a word is a stopword or not depends on the context of the
      text or of an application. In this case, the relevant task is
      detecting the sentiment of a tweet (positive, negative or neutral).
      In this task, the word `not' is often not considered a stopword, and
      it should be kept in the text. Please provide a version without
    stopwords of the following paragraph: `\{paragraph\}'. Print only the
  paragraph without stopwords, do not add any explanation, details or
notes.\\
  \midrule
    \multicolumn{1}{c}{\textbf{Lemmatization}}\\
    You specialize in text lemmatization. Text lemmatization is a
      natural language processing technique that is used to reduce words to
      their lemma, also known as the dictionary form. The process of
      lemmatization is used to normalize text and make it easier to
      process. For example, the verbs `is', `are', and `being' must all be
      reduced down to the common lemma 'be'. As another example, ``he's
      going'' must be lemmatized to ``he be go''. Lemmatization depends on
      correctly identifying the intended part of speech and meaning of a
      word in a sentence, as well as within the larger context surrounding
      that sentence, such as neighbouring sentences or even an entire
      document. Please provide the lemmatized version of this paragraph:
      `\{paragraph\}'. Print only the lemmatized paragraph, do not add any
    explanation, details or notes.\\
  \midrule
    \multicolumn{1}{c}{\textbf{Stemming}}\\
    You specialize in text stemming. Text stemming is a natural
      language processing technique that is used to reduce words to their
      base form, also known as the root form. The process of stemming is
      used to normalize text and make it easier to process. For example,
      the words `programming,' `programmer,' and `programs' can all be
      reduced down to the common stem `program'. As another example, the
      words `argue', `argued', `argument', `arguing', and `arguer' all stem
      to `argu'. Please provide the stemmed version of this paragraph:
      `\{paragraph\}'. Print only the stemmed paragraph, do not add any
    explanation, details or notes.\\
  \bottomrule
  \end{tabular}}
  \caption{Prompts used to lemmatize, stem and remove stopwords from the
    texts of the SemEval Sentiment dataset.}
  \label{tab:prompts}
\end{table*}

In this section, we describe the datasets, the evaluation metrics and the models
used to assess the effectiveness of
LLM-based preprocessing.

\paragraph{Datasets}

We select a suite of publicly available datasets encompassing binary and multiclass
classification tasks across multiple languages,
including
English, French, German, Italian, Portuguese and Spanish.
Specifically, for the evaluation of texts in English,
we use the Twitter datasets from SemEval-18 on emoji
prediction \cite{barbieri2018semeval} and irony detection
\cite{van2018semeval}, as well as from SemEval-19 on hate detection
\cite{basile2019semeval}, offensive language identification \cite{zampieri2019semeval}
and sentiment analysis \cite{nakov2013semeval}. In
addition, we evaluate
LLM-based preprocessing
on the task of News classification using
the Reuters \cite{lewis2004rcv1} and AG News \cite{NIPS2015_agnews}
datasets. Our main focus is on web-sourced data from platforms like
Twitter, which present unique challenges due to their informal language and
higher noise levels.
Unlike curated news
content, which tends to be cleaner and more uniform, Twitter data typically
require more extensive preprocessing to handle issues such as misspellings
and the presence of hashtags.
For %
non-English languages,
we employ five
datasets from the Tweet Sentiment Multilingual corpus \cite{barbieri-etal-2022-xlm}.
Due to high
computational costs, we randomly sample up to 3000 documents for training
and 3000 documents for evaluation while keeping the original class distributions.
Additionally, we create a
validation set of 2000 documents, extracted from the original SemEval-19
sentiment analysis training set.
These documents are used for tuning the hyperparameters of the ML
algorithms.

\paragraph{Models}
We compare five open source state-of-the-art LLMs,
encompassing different sizes and architectures:
Gemma-2-9B \cite{team2024gemma}, Gemma-3-4B \cite{team2025gemma}
LLama-3.1-8B \cite{dubey2024llama},
Phi-4-mini (3.8B parameters) \cite{abdin2024phi},
and
Qwen-2.5-7B \cite{qwen2024}
in their instruction-tuned version.
While \gemma-2 and \micphi have been primarily trained on English data, \gemma-3, \llama, \qwen are natively multilingual,
supporting Italian, Spanish, French, German and Portuguese.
We rely on the Hugging Face library to run the models, we set the
temperature to $0.7$ and, while generating texts, we use Sample Decoding
(i.e. \texttt{do\_sample=True}).
\cref{tab:prompts} provides examples of prompts used for lemmatization,
stemming, and stopword removal in English texts. These examples are based
on the SemEval-19 sentiment analysis dataset, with prompts for other datasets being either
straightforward adaptations or, in the case of multilingual datasets,
translations of those
shown here.

\paragraph{Baselines: traditional preprocessing}
We employ the stopword
lists and
stemmers
provided by \nltk, %
and the
lemmatizers provided by \spacy. %
The word ``not'' and language-specific negation lexicon are removed from
the \nltk's stopwords lists when preprocessing the SemEval and
Twitter Sentiment Multilingual datasets.

\paragraph{Machine Learning algorithms}%
  We use the scikit-learn %
  implementations of the Multinomial Naive Bayes \cite{webb2010naive},
  Decision Tree \cite{de2013decision}, and
  Logistic Regression \cite{nick2007logistic} algorithms.

\paragraph{Evaluation metrics}
  With respect to RQ1, for each preprocessing operation, we evaluate the accuracy
  of LLM-based preprocessing by computing the percentage of words in a text
  that are processed by the LLM in the same way as the corresponding
  traditional method.
Regarding RQ2, we use the micro $F_1$ measure for evaluating the performance of the
considered ML classification algorithms.

\begin{table*}[ht!]
  \centering
  \small
  \begin{tabular}{lccccccc}
    \textbf{Model} & \textbf{SW}    & \textbf{NSW}   & \textbf{L}     & \multicolumn{4}{c}{\textbf{S}} \\
                   &                &                &                & \textbf{Porter}                 & \textbf{Lanc.} & \textbf{Snow.} & \textbf{Any}   \\
    \toprule
    \gemma-2       & \textbf{84.29} & 13.95          & \textbf{82.61} & \textbf{74.93}                  & \textbf{63.15} & \textbf{75.65} & \textbf{81.14} \\
    \gemma-3       & 75.17          & 22.92          & 77.48          & 69.75                           & 57.46          & 71.53          & 74.51          \\
    \llama         & 81.61          & 30.04          & 79.20          & 66.54                           & 57.74          & 68.05          & 73.14          \\
    \micphi        & 43.53          & \textbf{12.74} & 78.56          & 61.61                           & 51.53          & 63.22          & 65.81          \\
    \qwen          & 79.37          & 22.15          & 82.37          & 61.11                           & 53.08          & 61.95          & 67.36          \\
    \bottomrule
\end{tabular}%
  \caption{%
    Accuracy of different LLMs in performing text preprocessing in English.
  }
  \label{tab:english_analysis}
\end{table*}

\begin{table*}[ht!]
  \centering
  \small
    \begin{tabular}{llcc|cc|cc|cc}
    \textbf{Language}                & \textbf{Model}                             &
    \multicolumn{2}{c}{\textbf{SW}}  &
    \multicolumn{2}{c}{\textbf{NSW}} &
    \multicolumn{2}{c}{\textbf{L}}   & \multicolumn{2}{c}{\textbf{S (Snowball)}} \\
    \toprule
    \multirow{5}{*}{French}
                                     & \gemma-2 & 96.83 & \textbf{97.94} & 28.02 & 33.12         & 61.06          & 53.28 & \textbf{51.51} & 34.47 \\
                                     & \gemma-3 & 95.01 & 68.46          & 34.18 & 25.30         & 63.93          & 62.64 & 50.91          & 34.28 \\
                                     & \llama   & 65.02 & 32.01          & 37.20 & 23.11         & 54.86          & 55.47 & 45.80          & 34.15 \\
                                     & \micphi  & 38.77 & 11.37          & 14.70 & \textbf{9.43} & 62.69          & 63.99 & 47.31          & 36.26 \\
                                     & \qwen    & 82.98 & 86.33          & 31.72 & 31.32         & \textbf{65.70} & 61.00 & 46.86          & 43.05 \\

            \hline
    \multirow{5}{*}{German}

                                     & \gemma-2 & 74.52 & \textbf{79.73} & 24.35 & 32.99          & 59.80 & 64.45          & \textbf{68.46} & 61.06 \\
                                     & \gemma-3 & 77.63 & 53.32          & 41.28 & 33.51          & 64.95 & \textbf{67.71} & 65.46          & 58.32 \\
                                     & \llama   & 58.08 & 25.65          & 35.44 & \textbf{16.54} & 58.38 & 60.07          & 58.54          & 57.32 \\
                                     & \micphi  & 25.74 & 30.33          & 17.46 & 20.84          & 57.24 & 58.90          & 50.84          & 45.30 \\
                                     & \qwen    & 57.23 & 47.04          & 31.29 & 22.88          & 64.84 & 64.29          & 53.17          & 48.86 \\

            \hline
    \multirow{5}{*}{Italian}

                                     & \gemma-2 & 86.48 & \textbf{89.20} & 20.82 & 22.36          & 59.30          & 58.31 & \textbf{56.65} & 50.86 \\
                                     & \gemma-3 & 86.45 & 86.70          & 31.20 & 26.98          & 63.37          & 61.66 & 46.29          & 48.80 \\
                                     & \llama   & 67.27 & 62.73          & 31.51 & 25.78          & 53.22          & 51.31 & 39.83          & 40.72 \\
                                     & \micphi  & 28.28 & 19.06          & 15.40 & \textbf{13.76} & 60.21          & 60.47 & 38.51          & 34.13 \\
                                     & \qwen    & 69.67 & 84.88          & 28.18 & 33.06          & \textbf{63.92} & 62.27 & 47.30          & 46.66 \\

                                     \hline
                                     \multirow{5}{*}{Portuguese}

                                       & \gemma-2 & 83.08          & 86.05 & 23.38          & 31.17 & 65.44          & 60.01 & 50.20 & \textbf{57.41} \\
                                       & \gemma-3 & \textbf{89.03} & 72.96 & 41.74          & 32.54 & 64.20          & 61.78 & 48.24 & 49.33          \\
                                       & \llama   & 65.00          & 72.52 & 23.94          & 36.58 & 61.80          & 62.16 & 44.04 & 47.80          \\
                                       & \micphi  & 34.75          & 49.13 & \textbf{15.10} & 19.11 & \textbf{70.42} & 68.92 & 42.02 & 36.70          \\
                                       & \qwen    & 68.30          & 76.09 & 25.06          & 32.79 & 70.17          & 69.02 & 49.50 & 46.74          \\
            \hline
    \multirow{5}{*}{Spanish}

                                     & \gemma-2 & 85.91 & \textbf{87.90} & 20.46 & 26.68          & 57.86          & 57.68 & \textbf{63.76} & 62.14 \\
                                     & \gemma-3 & 83.67 & 69.84          & 28.54 & 25.82          & 58.98          & 61.14 & 54.31          & 55.31 \\
                                     & \llama   & 69.99 & 33.53          & 29.02 & 21.44          & 51.02          & 52.11 & 48.69          & 51.90 \\
                                     & \micphi  & 28.42 & 18.66          & 19.06 & \textbf{14.94} & 55.26          & 55.02 & 45.86          & 44.24 \\
                                     & \qwen    & 67.46 & 75.55          & 26.84 & 25.81          & \textbf{62.30} & 61.11 & 54.78          & 52.05 \\
            \bottomrule
  \end{tabular}%
  \caption{%
    Accuracy of different LLMs in performing text preprocessing in five
    European languages.
    For each preprocessing operation, the values on the left and right
    refer to the scores obtained with an English prompt and with a
    language-specific prompt, respectively.}
\label{tab:multilingual_analysis}
\end{table*}

\paragraph{Hyperparameters settings}
To ensure a fair evaluation, we optimize the TF-IDF hyperparameters, such as the number of features and
the n-grams length,
on the Semeval-19 sentiment analysis validation set using
traditional preprocessing methods.
These optimized settings are consistently applied to both traditionally
processed and LLM-preprocessed text.

\section{Results}
\label{sec:results}

In \cref{sub:preprocessing_quality}, we examine the extent to which LLMs are able to replicate traditional preprocessing techniques.
As mentioned in the introduction, LLMs may %
identify different
stopwords, stems, and lemmas compared to traditional techniques, due to
their ability to manage contextual information.
We investigate whether this leads to improved performance in text
classification (RQ2) in \cref{sub:en_class}.

\subsection{LLMs' preprocessing abilities}%
\label{sub:preprocessing_quality}

\begin{table*}[ht!]
\centering
\small

    \tabcolsep=5pt
\begin{tabular}{llccccc}
\textbf{Dataset} & \textbf{Model} & \textbf{SW} &
\textbf{SW + L} & \textbf{L} &
\begin{tabular}[c]{@{}c@{}} \textbf{SW + S} \\ Porter | Lanc. | Snow. \end{tabular} & \begin{tabular}[c]{@{}c@{}} \textbf{S} \\ Porter | Lanc. | Snow. \end{tabular} \\ \toprule

\multirow{6}{*}{Emoji} %
& Traditional  & 21.15          & 21.41                         & 21.42          & 21.06 | 21.11 | 21.03 & 21.01 | 20.96 | 20.88 \\
& \gemma-2 & 21.52          & \texttt{ }\textbf{22.61\best} & \textbf{21.66} & 21.19                 & 21.00                 \\
& \gemma-3 & \textbf{22.00} & 22.17                         & 21.09          & \ul{21.28}            & \ul{21.27}            \\
& \llama   & 21.71          & 21.99                         & 21.22          & 20.51                 & 20.99                 \\
& \qwen    & 21.63          & \ul{22.53}                    & \ul{21.47}     & 20.97                 & 20.60                 \\
& \micphi  & \ul{21.73}     & 21.90                         & 21.46          & \textbf{21.58}        & \textbf{21.49}        \\
\midrule

\multirow{6}{*}{Hate}  %
& Traditional  & 48.73          & 48.93          & 49.67                         & 49.47 | 49.40 | 47.87 & 46.99 | 47.74 | 47.82 \\
& \gemma-2 & 49.61          & 47.47          & 49.52                         & 49.77                 & 49.24                 \\
& \gemma-3 & \textbf{50.87} & \textbf{50.68} & \ul{50.81}                    & \textbf{50.93}        & 49.49                 \\
& \llama   & 49.50          & \ul{49.75}     & \texttt{ }\textbf{51.31\best} & 50.45                 & 50.15                 \\
& \qwen    & \ul{50.80}     & 49.09          & 48.93                         & 49.61                 & \ul{50.34}            \\
& \micphi  & 49.38          & \ul{49.75}     & 49.91                         & \ul{50.67}            & \textbf{50.58}        \\
\midrule

\multirow{6}{*}{Irony} %
& Traditional  & 61.05                         & 60.11          & 59.73          & 61.96 | \textbf{63.01} | 61.39 & 60.96 | \ul{62.15} | 59.73 \\
& \gemma-2 & 61.64                         & \textbf{62.63} & 61.14          & 59.40                          & 60.63                      \\
& \gemma-3 & \ul{62.20}                    & 62.20          & 61.73          & 61.01                          & \textbf{62.88}             \\
& \llama   & 61.44                         & \ul{62.29}     & \textbf{63.35} & 59.31                          & 58.80                      \\
& \qwen    & 61.99                         & 61.22          & \ul{63.18}     & 57.95                          & 59.35                      \\
& \micphi  & \texttt{ }\textbf{64.50}\best & 59.14          & 61.82          & \ul{62.50}                     & 61.86                      \\
\midrule

\multirow{6}{*}{Offensive} %
& Traditional  & \ul{75.62}                    & \ul{74.53}     & \ul{73.19}     & \ul{75.73} | \textbf{75.93} | 74.61 & 74.22 | \ul{75.46} | \textbf{75.65} \\
& \gemma-2 & 74.81                         & \textbf{74.88} & 73.02          & 73.95                               & 72.71                               \\
& \gemma-3 & 73.37                         & 73.44          & 71.59          & 72.63                               & 71.98                               \\
& \llama   & \texttt{ }\textbf{76.71\best} & 73.95          & 71.47          & 73.76                               & 71.20                               \\
& \qwen    & 74.03                         & 73.84          & \textbf{74.38} & 72.40                               & 71.36                               \\
& \micphi  & 74.77                         & 74.26          & 72.09          & 73.57                               & 70.50                               \\
\midrule

\multirow{6}{*}{Sentiment} %
& Traditional  & \texttt{ }\textbf{48.89\best} & 48.05          & \textbf{48.71} & 47.81 | \textbf{48.54} | \ul{48.18} & 47.89 | \textbf{48.62} | \ul{48.61} \\
& \gemma-2 & \ul{48.13}                    & 47.77          & \ul{48.35}     & 46.59                               & 47.39                               \\
& \gemma-3 & 47.60                         & \textbf{48.80} & 47.27          & 43.64                               & 46.47                               \\
& \llama   & 47.96                         & \ul{48.13}     & 48.02          & 45.80                               & 47.06                               \\
& \qwen    & 46.98                         & 45.54          & 46.84          & 46.04                               & 46.38                               \\
& \micphi  & 47.24                         & 47.96          & 46.95          & 46.52                               & 45.31                               \\ \midrule

\multirow{6}{*}{AG News} %
& Traditional  & 61.44          & 62.21                         & 62.15          & 61.87 | 61.71 | 61.78 & 60.24 | \ul{61.64} | \textbf{61.78} \\
& \gemma-2 & \textbf{63.04} & \texttt{ }\textbf{66.04}\best & \ul{62.44}     & \textbf{62.66}        & 60.67                               \\
& \gemma-3 & \ul{62.89}     & \ul{63.69}                    & 61.34          & \ul{61.98}            & 59.37                               \\
& \llama   & 60.44          & 61.38                         & 60.37          & 56.79                 & 57.55                               \\
& \qwen    & 62.90          & 63.44                         & \textbf{62.69} & 58.16                 & 55.50                               \\
& \micphi  & 62.86          & 63.07                         & 60.45          & 60.33                 & 57.84                               \\ \midrule

  \multirow{6}{*}{Reuters} %
& Traditional  & \ul{85.76}     & 85.90                         & 86.40          & \textbf{85.83} | \ul{85.82} | 85.73 & 85.37 | \textbf{86.19} | 85.53 \\
& \gemma-2 & \textbf{86.00} & \texttt{ }\textbf{87.28}\best & \textbf{87.00} & 84.87                               & \ul{85.83}                     \\
& \gemma-3 & 83.73          & 86.60                         & \ul{86.66}     & 82.80                               & 85.78                          \\
& \llama   & 83.85          & 85.14                         & 84.86          & 81.56                               & 83.24                          \\
& \qwen    & 84.42          & \ul{86.74}                    & 85.85          & 80.11                               & 82.99                          \\
& \micphi  & 83.55          & 86.23                         & 86.41          & 82.30                               & 85.17                          \\

\bottomrule
\end{tabular}%
\caption{Comparison of LLM-based and traditional
  preprocessing on several
  text classification tasks. The scores are averages of the results
  obtained with three different ML algorithms. While applying
  traditional stemming, we
  report the values of the Porter | Lancaster | Snowball stemmers, following this order.}
\label{tab:results_classification_english}
\end{table*}

\newcommand{\mc}[1]{\multicolumn{1}{c}{#1}}
\newcommand{\mdc}[1]{\multicolumn{2}{c|}{#1}}
\newcommand{\mcoldc}[1]{\multicolumn{2}{c}{#1}}

\begin{table*}
  \centering
  \small
  \begin{tabular}{ll|cc|cc|cc|cc|cc|cc}
    \textbf{Dataset}   & \mc{\textbf{Model}}   & \mcoldc{\textbf{SW}} &
    \mcoldc{\textbf{SW + L}}
   & \mcoldc{\textbf{L}}  &
    \mcoldc{\textbf{SW + S}}
    & \mcoldc{\textbf{S}} \\ \toprule

  \multirow{5}{*}{French} %
  & Traditional  & \mdc{\textbf{52.95}} & \mdc{\textbf{52.49}} & \mdc{\ul{53.48}} & \mdc{\texttt{ }\textbf{53.98\best}} & \mcoldc{\textbf{52.87}} \\

  & \gemma-2 & 52.18                & \ul{52.53}           & 49.66            & 49.54                               & 51.88                    & 52.91          & \ul{51.65} & 49.54 & 47.81      & \mc{50.34} \\
  & \gemma-3 & 50.73                & 51.30                & 48.39            & 49.96                               & 51.49                    & 52.33          & 49.04      & 47.74 & 50.50      & \mc{48.70} \\
  & \llama   & 50.57                & 52.07                & 47.13            & 48.20                               & 50.19                    & 51.38          & 46.05      & 47.66 & 49.04      & \mc{48.00} \\
  & \micphi  & 51.69                & 52.26                & 48.08            & 50.04                               & 51.88                    & 52.50          & 51.03      & 49.62 & \ul{50.84} & \mc{50.15} \\
  & \qwen    & 50.50                & 49.89                & \ul{51.23}       & 49.80                               & 53.45                    & \textbf{53.52} & 50.08      & 47.16 & 47.85      & \mc{49.08} \\

                          \hline

  \multirow{5}{*}{German} %
  & Traditional  & \mdc{\texttt{ }\textbf{55.13\best}} & \mdc{\textbf{52.80}} & \mdc{53.18} & \mdc{\textbf{53.52}} & \mcoldc{\ul{54.06}} \\
  & \gemma-2 & 48.47                               & 49.77                & 50.04       & 46.93                & \ul{53.56}           & 50.26          & 50.27      & 49.31      & \textbf{54.48} & \mc{52.07} \\
  & \gemma-3 & 48.93                               & 49.46                & 46.93       & 47.55                & 53.14                & \textbf{53.79} & 48.51      & 48.74      & 53.64          & \mc{52.26} \\
  & \llama   & 47.13                               & 50.57                & 47.24       & 50.04                & 51.99                & 53.37          & 46.48      & \ul{50.46} & 50.96          & \mc{53.56} \\
  & \micphi  & 50.92                               & \ul{51.11}           & \ul{51.49}  & 49.20                & 52.64                & 53.45          & \ul{50.46} & 48.62      & 52.99          & \mc{51.07} \\
  & \qwen    & 46.59                               & 49.54                & 47.73       & 49.27                & 52.30                & 51.88          & 45.86      & 46.05      & 51.46          & \mc{48.35} \\
                          \hline

  \multirow{5}{*}{Italian} %
  & Traditional  & \mdc{\textbf{51.84}} & \mdc{\textbf{52.07}} & \mdc{50.61} & \mdc{\textbf{51.30}} & \mcoldc{52.33}                \\
  & \gemma-2 & 48.16                & 48.73                & 45.59       & 47.16                & 51.34                          & 50.61      & 45.51      & 44.80 & 52.30          & \mc{52.53}      \\
  & \gemma-3 & 46.74                & 46.90                & 44.48       & 46.25                & 52.03                          & 51.26      & 46.36      & 46.48 & \texttt{ }\textbf{53.68\best} & \mc{52.49}      \\
  & \llama   & 46.44                & 46.56                & 44.94       & 48.47                & 52.72                          & 51.46      & 42.34      & 43.87 & 48.62          & \mc{51.92}      \\
  & \micphi  & \ul{51.26}           & 49.00                & 48.62       & \ul{49.66}           & 52.49                          & 52.80      & \ul{51.12} & 49.54 & 52.26          & \mc{\ul{53.49}} \\
  & \qwen    & 45.48                & 44.71                & 45.94       & 41.88                & \textbf{53.37}  & \ul{52.99} & 43.44      & 44.48 & 48.97          & \mc{50.50}      \\
                           \hline

  \multirow{5}{*}{Portoguese} %
  & Traditional  & \mdc{\textbf{56.09}} & \mdc{\textbf{57.39}}          & \mdc{\textbf{57.16}} & \mdc{\textbf{56.51}} & \mcoldc{\texttt{ }\textbf{57.62}\best} \\
  & \gemma-2 & \ul{54.44}           & \textbf{56.09}                & 52.80                & 53.53                & 54.06                                   & 55.02      & 51.34 & 53.83      & 54.25 & \mc{\ul{54.94}} \\
  & \gemma-3 & 54.14                & \ul{54.44}                    & 53.10                & \ul{53.71}           & 54.33                                   & 54.94      & 53.64 & \ul{53.87} & 54.48 & \mc{54.48}      \\
  & \llama   & 53.26                & 52.53                         & 49.77                & 50.88                & 53.45                                   & 54.41      & 51.34 & 51.65      & 51.84 & \mc{52.34}      \\
  & \micphi  & 52.34                & 52.07                         & 51.92                & 50.19                & 53.98                                   & 53.64      & 48.54 & 50.22      & 50.04 & \mc{50.61}      \\
  & \qwen    & 52.37                & 53.49                         & 52.91                & 51.58                & \ul{55.17}                              & 53.03      & 52.03 & 47.70      & 51.00 & \mc{49.46}      \\
   \hline
  \multirow{5}{*}{Spanish} %
  & Traditional  & \mdc{47.47}          & \mdc{\textbf{49.43}}          & \mdc{48.47}          & \mdc{\textbf{49.88}} & \mcoldc{\textbf{49.61}}                \\
  & \gemma-2 & \ul{49.85}           & \texttt{ }\textbf{49.92}\best & 48.31                & 48.70                & 47.05                                   & 48.08      & 47.47 & \ul{48.74} & 48.43 & \mc{\ul{48.74}} \\
  & \gemma-3 & 47.36                & 47.62                         & 47.01                & 46.82                & \textbf{49.08}                          & 47.78      & 46.82 & 46.09      & 46.63 & \mc{47.32}      \\
  & \llama   & 48.40                & 48.62                         & 45.67                & 45.10                & 48.47                                   & 46.48      & 43.80 & 46.28      & 45.90 & \mc{47.55}      \\
  & \micphi  & 47.93                & 46.51                         & 46.25                & 44.67                & 46.44                                   & 47.51      & 47.28 & 46.32      & 44.71 & \mc{45.52}      \\
  & \qwen    & 49.39                & 47.70                         & \ul{49.08}           & 45.75                & 48.12                                   & \ul{48.74} & 46.67 & 45.33      & 46.86 & \mc{45.86}      \\

  \bottomrule

\end{tabular}%
\caption{Comparison of LLM-based and traditional preprocessing on the task
  of sentiment analysis in five European languages.
  The scores are averages of the results obtained with three different ML
  algorithms.
  For each combination of preprocessing operations, the value on the left
  refers to the score obtained with the English prompt, and the one on the right
refers to the score obtained with the language-specific one.
Traditional stemming is performed with the Snowball algorithm.
}
\label{table_results_classification}
\end{table*}

\cref{tab:english_analysis,tab:multilingual_analysis} compare
the preprocessing output produced by the LLMs with that produced by
traditional methods.
Specifically, \sw refers to the percentage of words
removed by the LLM
that match
\nltk's stopwords list, while \nsw
measures the percentage of words
removed by the LLM among those
that are \textit{not} considered stopwords by \nltk.
Additionally, \ls and \ss represent the percentage of words that are
respectively lemmatized and stemmed by the LLM exactly like the
corresponding traditional techniques.
For stemming in English (\cref{tab:english_analysis}),
the LLMs are first
compared against each of the Porter, Lancaster and Snowball algorithms,
then they are compared against the three algorithms collectively
(i.e., the LLM's output is valid if it matches the output of \textit{any}
of the three algorithms).
The reported values are averages over all texts in the same language.
These measures assess the similarity between LLM-based preprocessing
and traditional techniques, with the best-performing LLM being the one that
maximizes \sw, \ls and \ss, while minimizing \nsw. The best scores within
each dataset and preprocessing type are
highlighted in \textbf{bold} in \cref{tab:english_analysis,tab:multilingual_analysis}.

Since \gemma-2 is trained primarily on English data and it is the model with
the largest number of parameters, it would be expected to perform best on
English texts.
Indeed, %
\gemma-2 consistently outperforms all other models in stopword removal,
lemmatization, and stemming on English texts.
However, \micphi shows the most conservative behaviour in non-stopword
removal, achieving the lowest rate of non-stopwords removed compared to the
other LLMs.
Notably, this pattern is mostly consistent across all the analyzed
languages: \gemma-2 outperforms all the other LLMs in stopword
  removal and stemming (the only exception being \gemma-3 for stopword
  removal in Portuguese), while \micphi removes the fewest percentage of
non-stopwords.
  The lemmatization results, however, are not clear-cut: while \qwen
  achieves the best scores in French, Italian, and Spanish -- and performs
  comparably to \gemma-2 in English -- \gemma-3 and \micphi outperform the
  other models in German and Portuguese, respectively.
It is also interesting to highlight that  \gemma-3 -- despite being
  roughly half the size of \gemma-2, \llama, and \qwen{} -- consistently
  achieves performance comparable to, and occasionally surpassing, that of
the larger models.

Additionally, we note that language-specific prompts usually achieve
the highest \sw scores,
while the specification of the same prompts in English in most cases
produces the best lemmatization (\ls) and stemming (\ss) performance.
Overall, there is no consistent evidence that prompting in the target
language leads to better results. Specifically, \gemma-3 and \llama show
improved performance with language-specific prompts in 55\% and 60\% of
cases, respectively, whereas \gemma-2, \qwen, and \micphi perform better with
English prompts in 65\% of the cases analyzed.

We further observe that LLMs often eliminate words not traditionally
considered stopwords (\nsw column).
This behaviour supports our hypothesis that
LLMs' contextual
understanding influences stopword selection. For instance, ``user'' is
frequently removed, which is reasonable given that the datasets
include
social media text from Twitter \cite{barbieri-etal-2022-xlm}.
Regarding stemming, the lower overall scores compared to traditional
preprocessing may
be due to LLMs
generating different stems for the same word across different texts.
Although this deviates from traditional stemming rules, it might allow
for a more context-specific preprocessing, as also observed by
\cite{Sigir_stem}.

Overall, these results show that LLMs are quite effective at identifying
stopwords across multiple languages, with \gemma-2 detecting more than $97\%$ of
stopwords in
French texts, and at least $79\%$ in other languages. Additionally,
they show strong
lemmatization capabilities in English, with all the evaluated models
correctly identifying over $77\%$ of lemmas.

\subsection{Text classification}
\label{sub:en_class}

\cref{tab:results_classification_english,table_results_classification} report
the average performance score of the three ML models that we have
trained on English (\cref{tab:results_classification_english}) and
non-English texts (\cref{table_results_classification}), by applying the
LLM-based preprocessing and traditional preprocessing methods.
Each column corresponds to a specific preprocessing task: \sw denotes
stopword removal, \swl applies lemmatization followed by stopword removal,
\ls represents lemmatization alone, \sws combines stopword removal and
stemming, and \ss applies stemming only.
For each dataset and preprocessing task, the best results are highlighted
in \textbf{bold}, while the second-best scores are \ul{underlined}.
The best result within each dataset is marked with a $\bestinline$.

\paragraph{English language}
We first note that LLMs
outperform traditional methods
in all datasets except for
Sentiment,
and more specifically
in 25 out of the 35 examined combinations of datasets and preprocessing tasks.
Moreover, in 80\% of these cases, the second-best result is also achieved by an LLM.
Traditional preprocessing outperforms LLM-based preprocessing in 10 out of
35 combinations,
with a margin greater than $1$ point in $F_1$ in 5 of them (Offensive
SW+S and S, Sentiment SW+S and S, and AG News S).
Notably, LLMs achieve the highest performance in stopword removal
  combined with lemmatization in all the proposed datasets, indicating
  their ability to dynamically identify
task-relevant stopwords and lemmas in a more context-sensitive manner than
traditional techniques. In particular, \gemma-2 achieves a 6.16\%
improvement over traditional techniques in the AG News dataset.
Additionally, LLM-based preprocessing outperforms traditional
stopword removal and lemmatization in 6 out of 7 datasets. The only
exception is the Sentiment dataset, where \gemma-2 shows
a slight
underperformance.

Our results indicate however that stemming with LLMs is not as effective as
other preprocessing operations: indeed LLM-based preprocessing
outperforms traditional stemming in only 3 out of the 7 datasets. Several factors
may contribute to this outcome.
First, stemming is a task where context plays a limited role, making it
less sensitive to the contextual capabilities of LLMs. This aligns with
findings by \cite{Sigir_stem}, who show that LLMs' stemming performance is
suboptimal in an information retrieval pipeline. Furthermore, we note that
LLMs exhibit inconsistencies in stemming across
documents. Unlike
traditional algorithms, which apply fixed rules, LLMs may stem the same
word differently depending on context. For instance, in some cases, an LLM
may generate a stem that matches the Porter stemmer, while in others, it may
align with the Lancaster stemmer or be completely different. This lack of consistency
results in non-standardized text representations, which can negatively
impact downstream tasks such as lexical
feature extraction, and consequently
their classification performance.

\paragraph{Non-English languages}

\cref{table_results_classification} presents the performance of text
classification across French, German, Italian, Portuguese and Spanish.
Overall, LLMs achieve performance on par (within 1 point) with or even better than traditional techniques in half of the evaluated cases. %
Notably, LLMs achieve the highest performance in 4 out of 5 datasets when lemmatization is applied, showing their ability to understand contextual information even in non-English languages.
Moreover, LLMs achieve the highest performance in the Italian and Spanish datasets (marked with $\bestinline$) and perform only marginally lower than the best score in French and German.

Interestingly, \gemma-2 and -3 outperform traditional methods even in
stemming for Italian and German, respectively.
This finding contrasts with the one observed in the English setting.
For Spanish, \gemma-2 also shows a significant improvement in stopword removal, underscoring its ability to identify stopwords based on context.
Moreover, in Spanish, traditional preprocessing
outperforms LLMs by only 1 point in stemming,
and similarly in stopword removal when combined with lemmatization or stemming.

Notably, the performance of \llama improves when using
language-specific prompts in 80\% of the preprocessing tasks.
For both \gemma-2 and \gemma-3 
this is
instead true in more than 60\% of the
analyzed combinations,
and for \qwen and \micphi
only in 40\%.
This finding is unexpected, given that \qwen is inherently multilingual,
while \gemma-2 is mostly trained on English data.

\section{Limitations}%
\label{sec:limitations}

Regarding RQ1,
the ability of LLMs to perform text preprocessing is evaluated by comparing
their outputs to those generated by well-known Python libraries, such as
\nltk and \spacy. There may however be instances where LLMs outperform
these libraries -- for example, by splitting a long hashtag like
``\texttt{\#illegalaliens}'' and correctly lemmatizing it as
``\texttt{illegal alien}'' -- that are not accounted for in the evaluation
metrics.
We do not perform extensive prompt engineering in this work, as we are
interested in investigating the abilities and raw behaviour of Large
Language Models rather than obtaining the best results. This is also due to
computational constraints and costs. However, some results may differ if
other prompts are considered.
Another limitation of
LLM-based preprocessing
is the high computational cost
of using LLMs,
which is significantly greater than
that of traditional methods. Therefore, LLM-based preprocessing
is best justified for low-resource languages, such as languages that
  lack the extensive amounts of annotated resources that are needed to
develop or train lemmatizers.
Our results, which show that LLMs can consistently match or even
surpass traditional preprocessing techniques across multiple
languages, further support their use in such contexts.

\section{Conclusions and Future Works}%
\label{sec:conclusions_and_further_research}

In this paper, we investigate the capability of LLMs to perform text
preprocessing, including stopword removal, lemmatization, and stemming.
We conduct a comparative analysis of various LLMs, differing in both size
and architecture, to assess their ability to replicate traditional
preprocessing techniques across five languages. Additionally, we evaluate
the impact of LLM-based preprocessing on multiple downstream Machine
Learning classification tasks by training models on text preprocessed using
traditional and LLM-based approaches.
Our findings indicate that LLMs outperform
traditional lemmatization techniques across most of the evaluated
languages and datasets, and consistently improve stopword removal in
English, both with and without lemmatization. However, LLMs do not appear
to perform competitively in stemming.
Although preprocessing for European languages has been extensively studied,
we note that stemmers and lemmatizers for many other languages have
received significantly less attention \cite{silvello2018statistical}, often
resulting in reduced effectiveness. Given the promising results achieved,
future work will explore the potential of LLMs as stemming and
lemmatization tools for low-resource languages.

\section*{Acknowledgment}
We acknowledge the CINECA award under the ISCRA initiative, for the availability of high-performance computing resources and support.

\printbibliography[heading=bibintoc]

\end{document}